\documentclass[10pt,twocolumn,letterpaper]{article}

\usepackage{iccv}
\usepackage{times}
\usepackage{epsfig}
\usepackage{graphicx}
\usepackage{amsmath}
\usepackage{amssymb}

\usepackage{booktabs}
\usepackage{multirow}
\usepackage{threeparttable}
\usepackage{pgfplots}
\usepackage{colortbl}
\pgfplotsset{compat=newest,compat/show suggested version=false}
\newcommand{\gray}[1]{\textcolor[RGB]{140,140,140}{#1}}
\usepackage{makecell}
\usepackage{pifont}
\usepackage{cite}

\usepackage[pagebackref=true,breaklinks=true,letterpaper=true,colorlinks,bookmarks=false]{hyperref}

\iccvfinalcopy 

\def\iccvPaperID{9567} 

\ificcvfinal\fi

\begin{document}


\title{Masked Spatio-Temporal Structure Prediction for Self-supervised Learning \\ on Point Cloud Videos}

\author{
Zhiqiang Shen$^{1,5}$\footnotemark[1]\ , \ \ 
Xiaoxiao Sheng$^{1}$\footnotemark[1]\ , \ \ 
Hehe Fan$^{2}$\footnotemark[2]\ , \ \ 
Longguang Wang$^{3}$, \ \ 
Yulan Guo$^{4}$, \ \ \\ 
Qiong Liu$^{5}$, \ \ \ \ 
Hao Wen$^{5}$, \ \ \ \ 
Xi Zhou$^{1,5}$ \ \ \ \\
$^1$Shanghai Jiao Tong University\ \ \ \ \ 
$^2$Zhejiang University\ \ \ \ \ 
$^3$Aviation University of Air Force\\
$^4$Sun Yat-sen University\ \ \ \ \ \ \ \ \ \ \ 
$^5$CloudWalk\\
}
\maketitle

\renewcommand{\thefootnote}{\fnsymbol{footnote}}
\footnotetext[1]{These authors contributed equally.}
\footnotetext[2]{Corresponding author.}

\ificcvfinal\thispagestyle{empty}\fi

\begin{abstract}

Recently, the community has made tremendous progress in developing effective methods for point cloud video understanding that learn from massive amounts of labeled data.
However, annotating point cloud videos is usually notoriously expensive. 
Moreover, training via one or only a few traditional tasks (\eg, classification) may be insufficient to learn subtle details of the spatio-temporal structure existing in point cloud videos.
In this paper, we propose a Masked Spatio-Temporal Structure Prediction (MaST-Pre) method to capture the structure of point cloud videos without human annotations. 
MaST-Pre is based on spatio-temporal point-tube masking and consists of two self-supervised learning tasks. 
First, by reconstructing masked point tubes, our method is able to capture the appearance information of point cloud videos. 
Second, to learn motion, we propose a temporal cardinality difference prediction task that estimates the change in the number of points within a point tube. 
In this way, MaST-Pre is forced to model the spatial and temporal structure in point cloud videos. 
Extensive experiments on MSRAction-3D, NTU-RGBD, NvGesture, and SHREC’17 demonstrate the effectiveness of the proposed method. 
The code is available at \url{https://github.com/JohnsonSign/MaST-Pre}. 
\end{abstract}

\section{Introduction}
\label{sec:Introduction}
In physics, motion is the phenomenon in which position changes over time. 
Because point clouds provide precise position information, \ie, 3D coordinates, point cloud videos, which evolve over time, can accurately describe the 3D motion in the real world. 
Effectively understanding point cloud videos can significantly improve intelligent agents on the interaction with environments. 
Therefore, the community has developed a few effective methods for point cloud video understanding, including video classification~\cite{pstnet, p4d, fan2022point, fan2021deep, 3dv, zhong2022no} and semantic segmentation~\cite{PST2, wen2022point, MeteorNet, minke}. 
However, most of these methods are based on supervised learning and that requires much effort to carefully annotate massive amounts of labels. 
Moreover, learning via only classification or segmentation may make deep neural networks take too much emphasis on the task itself but largely ignore the subtle details of the instinct spatio-temporal structure in point cloud videos. 
To alleviate those problems, we propose a self-supervised learning method on point cloud videos.  

    \begin{figure}[t]
        \centering
            \includegraphics[width=1\linewidth]{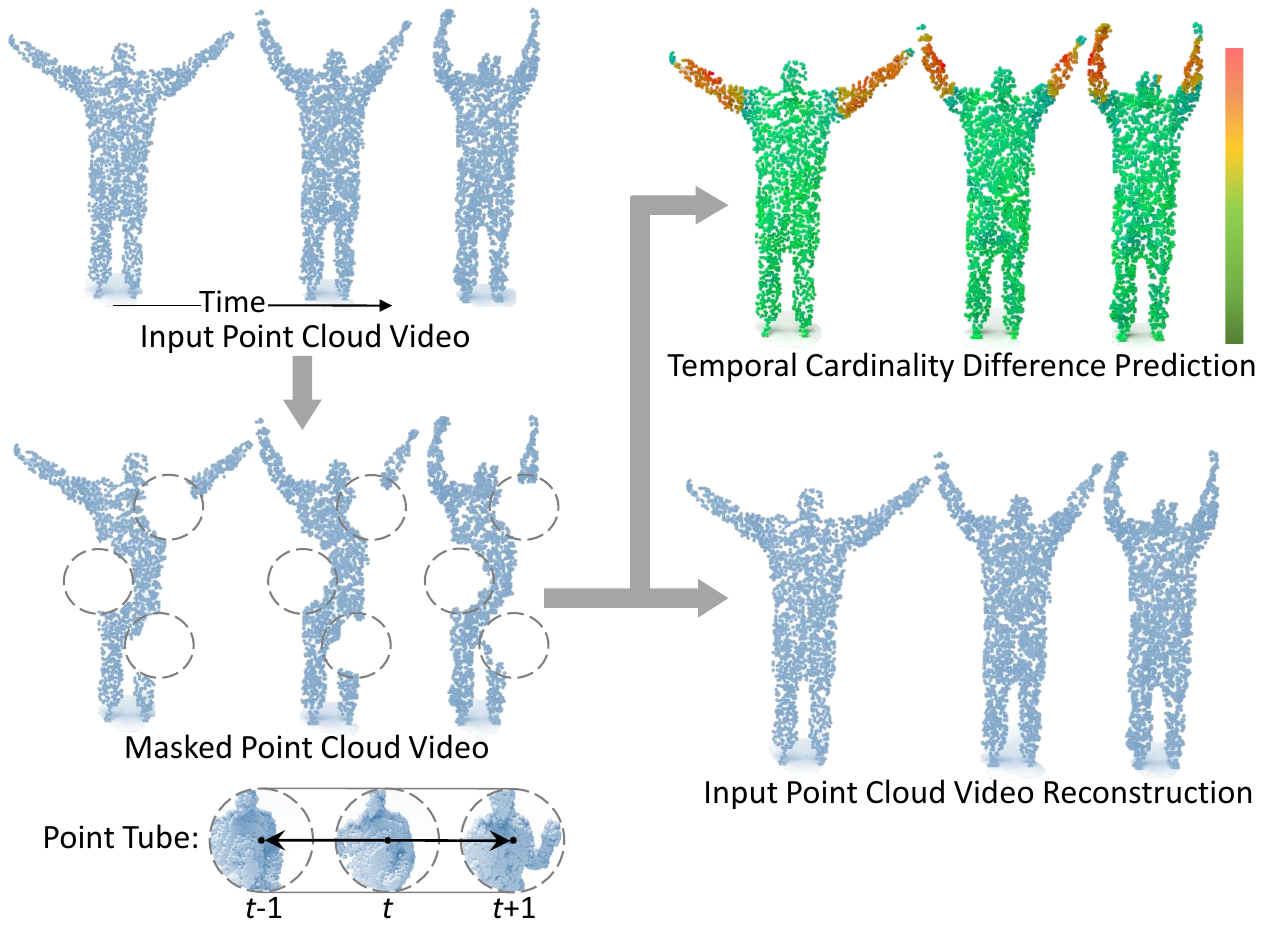}
            \caption{
            Our MaST-Pre is based on spatio-temporal point-tube masking.
            To enable a model to capture the appearance structure in point cloud videos, we ask it to reconstruct masked point tubes. 
            To equip the model with motion modeling ability, we develop a temporal cardinality difference prediction task. 
            }
        \label{fig1}
    \end{figure}

Self-supervised learning uses supervisory signals from the data itself and enables deep neural networks to learn from massive data without human annotations. This is important to recognize more subtle patterns in data.  
Networks pre-trained with self-supervised learning usually yield higher performance than when solely trained in a supervised manner \cite{kenton2019bert, he2022masked, chen2020simple, he2020momentum}.  
Although self-supervised learning has been applied to images \cite{he2022masked}, videos \cite{feichtenhofer2022masked, tong2022videomae} and static point clouds \cite{pang2022masked, yu2022point}, it has not been promoted on 4D signals, such as point cloud videos. 
Visual signals in point cloud videos can be divided into appearance and motion. 
While appearance specifies which objects are in videos, motion describes their dynamics. 
Therefore, self-supervised learning on point cloud videos should carefully make the most of the appearance and motion structure. 

In this paper, we propose a Masked Spatio-Temporal Structure Prediction (MaST-Pre) method for self-supervised learning on point cloud videos (Fig.~\ref{fig1}).  
MaST-Pre is based on a masking strategy, which has been proven effective in a range of applications. 
For example, because of the canonical structure, images can be easily segmented into multiple patches for masking \cite{dosovitskiy2020image, he2022masked}, which in the case of video are extended to patch tubes \cite{feichtenhofer2022masked, tong2022videomae}. 
For unstructured static point clouds, spherical support domain masking can be used for masked autoencoder~\cite{pang2022masked, yu2022point}. 
However, the spatial irregularity and temporal regularity make point cloud videos require a more elaborate masking strategy. 
Our method is based on a masked point tube mechanism, where a point tube is a local area expanding over a short time \cite{pstnet}.

Based on point-tube masking, our MaST-Pre employs two self-supervised tasks to capture the appearance and motion structure, respectively. 
First, to learn the appearance structure, MaST-Pre is asked to predict the invisible parts of the input from unmasked points. 
Second, to capture the dynamics in point tubes, we propose the \emph{temporal cardinality difference}, which can be calculated online from inputs without additional parameters. 
Cardinality can reflect basic structures (\eg, line, edge, and plane) of static point clouds~\cite{LiangAdaptationECCV2022}. 
In this paper, we extend it to a temporal version so that it can model the dynamics of point cloud videos. 
Intuitively, the temporal cardinality difference characterizes the flow of points within a short time. 
Therefore, inferring the temporal cardinality difference of masked point tubes facilitates MaST-Pre to learn motion-informative representations. Our contributions are summarized as follows:

\begin{itemize}
\item We design a 4D scheme of masked prediction for self-supervised learning on point cloud videos, termed as MaST-Pre. 
Our MaST-Pre jointly learns the appearance and motion structure of point cloud videos.

\item We propose the temporal cardinality difference, a simple and effective motion feature directly captured from raw input points. It explicitly guides MaST-Pre to learn motion-informative representations.

\item Extensive experiments and ablation studies on several benchmark datasets validate that our MaST-Pre learns rich representations of point cloud videos.
\end{itemize}

\begin{figure*}[ht]
	\centering
	\includegraphics[width=1\linewidth]{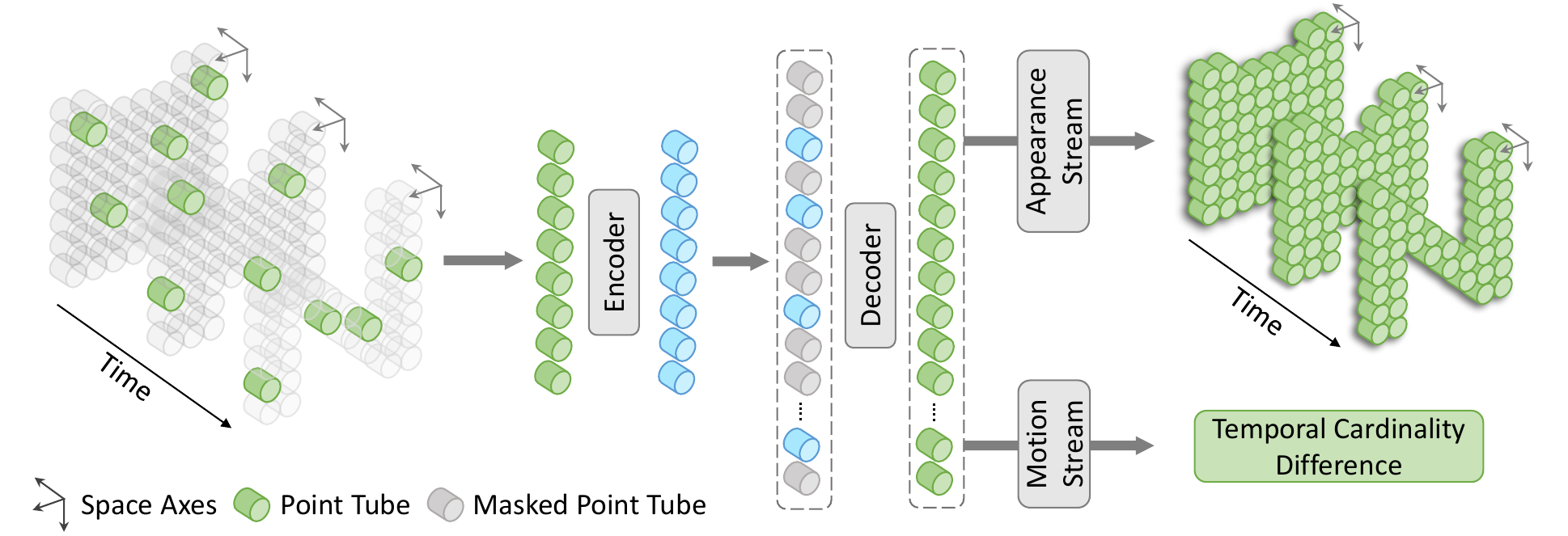}
	\caption{Illustration of the proposed MaST-Pre method. First, given a point cloud video, MaST-Pre divides it into several point tubes and masks part of them. Then, based on an encoder-decoder architecture, MaST-Pre attempts to recover masked point tubes and predict their temporal cardinality difference.}
	\label{fig2}
\end{figure*}

\section{Related Work}
\label{sec:Related Work}
In this section, we first briefly review visual mask prediction for self-supervised learning. Then, we present recent advances in self-supervised learning on point clouds and dynamics modeling for point cloud videos.

\subsection{Visual Mask Prediction} 
Mask prediction has been proven to be an excellent self-supervised task for visual representation learning \cite{dosovitskiy2020image, xie2022simmim, he2022masked}. By reconstructing target signals from the masked input, mask prediction enables the network to learn rich representations and boosts self-supervised learning \cite{iGPT, bao2021beit, wang2022bevt, MaskFeat}. Chen~\etal~\cite{iGPT} extended GPT~\cite{GPT} to operate the pixel sequence for prediction. Bao~\etal~\cite{bao2021beit} and Wang~\etal~\cite{wang2022bevt} introduced another successful framework, BERT~\cite{kenton2019bert}, to predict the identities of masked tokens.

Then, He~\etal~\cite{he2022masked} proposed MAE as a scalable vision learner to predict the pixels of masked patches. Feichtenhofer~\etal~\cite{feichtenhofer2022masked} and Tong~\etal~\cite{tong2022videomae} extended MAE to video representation learning by masking patch tubes. Wei \etal \cite{MaskFeat} developed MaskFeat to predict the HOG features of masked spatio-temporal tubes for self-supervised video pre-training. The regular structure of images and videos makes it easy to obtain patches or patch tubes, which facilitates the design of masking strategies. Pang~\etal~\cite{pang2022masked} extended MAE to unstructured static point clouds and designed a masking strategy based on the local spatial neighborhoods. However, point cloud videos are not only spatially irregular but also temporally misaligned across frames \cite{p4d, pstnet}. To remedy this, we design a point-tube masking strategy.

\subsection{Self-supervised Learning on Point Clouds}
Contrastive learning has made significant progress on static point clouds \cite{xie2020pointcontrast, hou2021exploring, zhang2021self, rao2020global}. Xie~\etal~\cite{xie2020pointcontrast} proposed PointContrast to discriminate two geometric views of matched points using contrastive loss. Hou \etal \cite{hou2021exploring} introduced contextual contrastive learning to PointContrast for data-efficient point pre-training. Rao \etal \cite{rao2020global} mapped the local and global features to shared representation space and applied a contrastive loss on them. Zhang \etal \cite{zhang2021self} used the instance discrimination task on two augmented versions of a point cloud, while Huang \etal \cite{huang2021spatio} pre-trained static point clouds using spatio-temporal augmentations. They transformed different views at point, region, object, and scene levels, and then used contrastive learning to judge their semantic consistency. However, there are limited augmentation methods that can guarantee the semantic consistency of point clouds, let alone point cloud videos.

Prediction-based methods on point clouds also attract a lot of attention. Yang \etal \cite{yang2018foldingnet} proposed a folding-based autoencoder that deforms a 2D grid to reconstruct the target 3D point cloud. Recently, mask prediction has been extended to static point clouds. Liu~\etal~\cite{liu2022masked} designed a point discrimination task for masked patches. Yu~\etal~\cite{yu2022point} proposed PointBERT with an offline point tokenizer. Pang~\etal~\cite{pang2022masked} used a simpler task, reconstructing masked point coordinates, for point pre-training in an end-to-end manner. However, these methods solely focus on the geometric representation learning of static point clouds. In this paper, on top of learning appearance information, an explicit motion information learning method is designed for point cloud videos.

\subsection{Dynamics Modeling for Point Cloud Videos}
Supervised learning dominates point cloud video research. \cite{pstnet, fan2021deep} and \cite{p4d, fan2022point, PST2, wen2022point} use convolution-based methods and attention-based methods to implicitly learn the long-term features of point cloud videos, respectively. In addition, \cite{3dv, zhong2022no, MeteorNet} apply empirical-based dynamics methods modeling point cloud videos. Wang \etal \cite{3dv} introduced temporal rank pooling \cite{fernando2016rank} to capture the frame-level dynamics. Zhong \etal \cite{zhong2022no} proposed a two-stream framework and used feature-level ST-surfaces \cite{pottmann2001computational} in the dynamics learning branch. Liu \etal \cite{MeteorNet} used a scene flow estimator \cite{liu2019flownet3d} or an alternative grouping method to do point tracking for dynamics modeling. Although these methods are effective, they require complex calculations or additional modules.

Self-supervised learning on point cloud videos is understudied \cite{Shen_2023_CVPR, Sheng_2023_AAAI}. Wang~\etal~\cite{wang2021self} pre-trained the encoder by predicting the temporal order of shuffled segments to learn the dynamics of point cloud videos. Zhang \etal \cite{zhang2022completetopartial} developed complete-to-partial 4D distillation to predict the representations of point cloud frames within a short time window. However, these methods learn motion information using clip-level or frame-level pretext tasks. In this paper, we propose the temporal cardinality difference prediction for fine-grained dynamics learning on point cloud videos.

    \begin{figure*}[ht]
        \centering
        \includegraphics[width=1\linewidth]{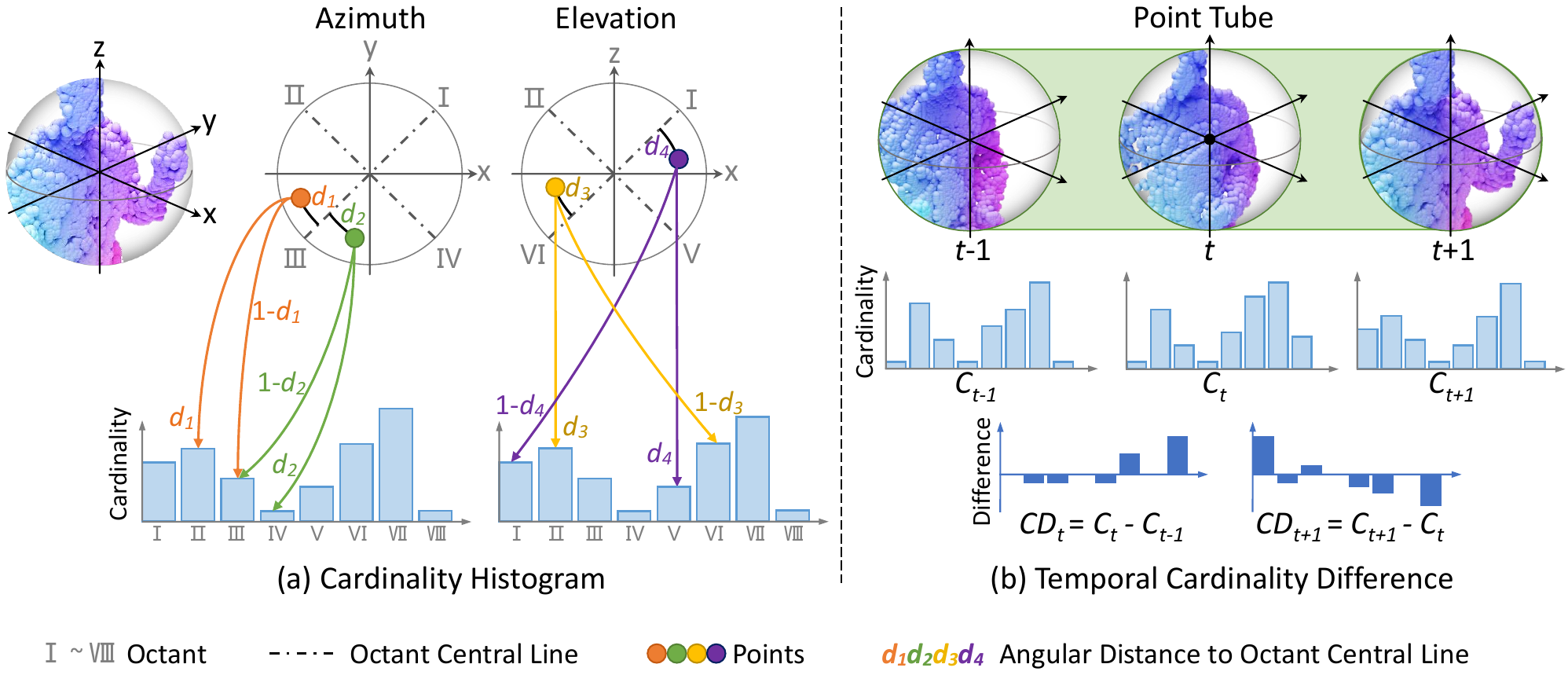}
        \caption{Illustration of Cardinality Histogram (a) and Temporal Cardinality Difference (b).}
        \label{fig3}
    \end{figure*}
    
\section{Method}
The architecture of our MaST-Pre is illustrated in Fig.~\ref{fig2}. 
Given a point cloud video, it is first divided into multiple point tubes.
Next, a masking operation is performed, separating these point tubes into visible and invisible parts.
The visible ones are fed to an encoder, and then their updated embeddings are fed to the decoder along with the masked point tubes.
Afterward, two-stream prediction tasks are implemented to recover the point coordinates within masked point tubes and to infer their temporal cardinality differences, respectively.
Intuitively, enabling the decoder to perform well on two-stream prediction tasks demands the encoder to learn representations rich in appearance and motion information jointly.

\subsection{Masking Strategy}
\label{subsec:Masking Strategy}
The masking strategy includes three steps: input division, embedding, and masking operation.

\textbf{Division.}
Point tubes are introduced as division units of the input point cloud video.
Specifically, given a point cloud video $\boldsymbol{P}$, Farthest Point Sampling is used to select $N$ key points $\boldsymbol{\hat{p}}$ from the input. 
Next, we construct one point tube for each key point. The point tube centered at $i$-th key point $\hat{p}_i$ is denoted as $\boldsymbol{Tube}_{\hat{p}_i}\!=\!\{p\ |\ p\!\in\!\boldsymbol{P},\ \mathcal{D}_{s}(p,{\hat{p}_i})\!<\!r,\ \mathcal{D}_{t}(p,{\hat{p}_i})\!<\!\frac{l}{2}\}$, where $p$ is one of the input points, $\mathcal{D}_{s}$ is the Euclidean distance, $\mathcal{D}_{t}$ is the difference in frame timestamps of two points, $r$ is the radius of a spatial neighborhood and $l$ is the number of frames in a point tube. 
Then, we use random sampling to select $n$ points in each spatial neighborhood.

In this way, a point cloud video is divided into $N$ point tubes, and each point tube contains $l \times n$ points. To ensure all points are covered by point tubes, $r$ and $l$ are set to maintain a minor overlap between adjacent point tubes.

\textbf{Embedding.}
Following our baseline \cite{p4d}, each point tube is encoded into an embedding:
    \begin{equation}
        \centering        
        \boldsymbol{E}_{\hat{p}_i} = \sum_{t=1}^{l}\ {\sum_{p\in{{\boldsymbol{Tube}_{\hat{p}_i}^{t}}}}{f(p-{\hat{p}_i})}},
    \end{equation}
where $\boldsymbol{Tube}_{\hat{p}_i}^{t}$ is the $t$-th frame of the point tube centered at $\hat{p}_i$ and $f(\cdot)$ is an MLP-based feature extractor. More details can be found in \cite{p4d}.

\textbf{Masking.}
The design of the masking operation is related to the information redundancy of input \cite{he2022masked, feichtenhofer2022masked, tong2022videomae}. 
Because of spatio-temporal coherence, the redundancy of the point cloud video is higher than low-dimensional data. Therefore, our MaST-Pre uses a high masking ratio on point cloud videos and the empirical result is 75\%. 
A high ratio is helpful to alleviate information leakage and make reconstruction a meaningful self-supervised task.
In addition, our MaST-Pre uses random masking of point tubes. As a spacetime-agnostic method, random masking is more effective than structure-aware strategies \cite{feichtenhofer2022masked}.

\subsection{Autoencoding} 
\label{subsec:Autoencoding}
Our autoencoder is based on vanilla Transformers of point cloud videos \cite{p4d}. An asymmetric encoder-decoder design is introduced to MaST-Pre.

\textbf{Encoder.}
To better capture the dynamics of point cloud videos, joint spatio-temporal attention is adopted \cite{arnab2021vivit, liu2022video}.
In addition, only the visible point tubes with spatio-temporal positional embeddings are fed into the encoder during pre-training.

\textbf{Decoder.}
The decoder is similar to our encoder but a lightweight vanilla Transformer \cite{p4d}.
It takes both encoded point tubes and masked ones as input. 
By adding a full set of spatio-temporal positional embeddings to all tokens, location clues are provided for self-supervised learning. 
After decoding, only the embeddings of masked point tubes are fed to the following prediction heads.

\subsection{Two-stream Prediction}
\label{subsec:two-stream prediction}
It is demonstrated in \cite{qian2022static, NIPS2014_00ec53c4} that effective video representation integrates appearance and motion information.
Therefore, two-stream self-supervised tasks are proposed to explicitly predict motions and reconstruct the appearance of masked point cloud videos.

\textbf{Appearance Stream.}
The reconstruction objectives are the point coordinates of masked point tubes. 
The $l_2$ Chamfer Distance loss is used between predictions $\boldsymbol{P}_{pre} \in \mathbb{R}^{{l}\times{n}\times{3}}$ and the ground truth $\boldsymbol{P}_{gt}\in \mathbb{R}^{{l}\times{n}\times{3}}$:
    \begin{equation}
    \centering
        \begin{aligned}
        \mathcal{L}_{app}=\frac{1}{l}\sum_{i=1}^{l}
        &\left \{ \frac{1}{|\boldsymbol{P}_{pre}^{i}|}\sum_{a \in \boldsymbol{P}_{pre}^{i}}{\min_{b \in \boldsymbol{P}_{gt}^{i}}}{\Vert{a-b}\Vert}_2^2 + \right. \\
        &\left. \frac{1}{|\boldsymbol{P}_{gt}^{i}|}\sum_{b \in \boldsymbol{P}_{gt}^{i}}{\min_{a \in \boldsymbol{P}_{pre}^{i}}}{\Vert{b-a}\Vert}_2^2 \right \}.
        \end{aligned}
    \label{app-loss}
    \end{equation}

\textbf{Motion Stream.}
We propose the \textit{temporal cardinality difference} as the target of the motion prediction stream.
As shown in Fig.~\ref{fig3}(a), the spherical support domain of the key point is divided into 8 octants. 
We follow the conventional rule that the area where the $xyz$-coordinates are greater than 0 is the first octant \gray{\uppercase\expandafter{\romannumeral1}} and then increases counterclockwise.

Then, count the cardinality of each octant into the corresponding bin of a histogram.
To alleviate the noise in real-world point clouds, a probabilistic approach is employed. 
Specifically, calculate the angular distance $d$ of each point to the central line of its current octant, and divide it by $90^{\circ}$ to normalize. 
For example, the current octant of \textbf{Point2} (the green one in Fig.~\ref{fig3}(a)) is octant \gray{\uppercase\expandafter{\romannumeral3}}. The angular difference $d2$ between \textbf{Point2} and the central dashed line of octant \gray{\uppercase\expandafter{\romannumeral3}} is $30^{\circ}$. Consequently, the probabilities of \textbf{Point2} belonging to octant \gray{\uppercase\expandafter{\romannumeral4}} and  \gray{\uppercase\expandafter{\romannumeral3}} are $\frac{1}{3}$ (\ie, $\frac{30^{\circ}}{90^{\circ}}$) and $\frac{2}{3}$ (\ie, $\frac{60^{\circ}}{90^{\circ}}$), respectively. 
In particular, when a point falls on an axis (\eg, the $-y$ axis), its probabilities belonging to octant \gray{\uppercase\expandafter{\romannumeral4}} and \gray{\uppercase\expandafter{\romannumeral3}} are both 0.5.

Next, as shown in Fig.~\ref{fig3}(b), the cardinality histograms $\boldsymbol{C}\!\in\!\mathbb{R}^8$ between adjacent frames of a point tube are subtracted to obtain its temporal cardinality difference $\boldsymbol{C\!D}\!\in\!\mathbb{R}^8$, which constitutes the ground truth $\boldsymbol{M}_{gt}\!\in\!\mathbb{R}^{{(l\!-\!1)}\times 8}$ of the motion stream.
The decoded embeddings of masked point tubes are passed through a linear layer to obtain the motion prediction $\boldsymbol{M}_{pre}\!\in\!\mathbb{R}^{{(l\!-\!1)}\times 8}$.
The smooth $l_1$ loss of $i$-th $\boldsymbol{C\!D}$ between prediction and ground truth is denoted as $\mathcal{L}_{m}^{i}$. The loss of our motion stream is denoted as $\mathcal{L}_{motion}$:
    \begin{equation}
    \centering
        \mathcal{L}_{m}^{i}\!=\!\begin{cases}  
        0.5\!\times\!\!({\boldsymbol{M}_{pre}^{i}\!\!-\!\boldsymbol{M}_{gt}^{i}})^2, &\!\text{if}\ |{\boldsymbol{M}_{pre}^{i}\!\!-\!\boldsymbol{M}_{gt}^{i}}|\!<\!1 \\
        |{\boldsymbol{M}_{pre}^{i}\!\!-\!\boldsymbol{M}_{gt}^{i}}|-0.5, &\!\text{otherwise} \\
        \end{cases},
    \end{equation}
    \begin{equation}
    \centering
        \mathcal{L}_{motion}\!=\!\frac{1}{l-1} \sum_{i=1}^{l-1}{\mathcal{L}_{m}^{i}}.
    \end{equation}

Overall, the total loss of our MaST-Pre is defined as:
    \begin{equation}
        \centering        
        \mathcal{L}_{total}=\mathcal{L}_{app}+\mathcal{L}_{motion}.
    \end{equation}
With both loss terms, our MaST-Pre can simultaneously learn the geometry and dynamics of point cloud videos.

\section{Experiments}
Our experiments are conducted on four point cloud video datasets. Following \cite{he2022masked, feichtenhofer2022masked, tong2022videomae}, we implement end-to-end fine-tuning, semi-supervised learning, and transfer learning to evaluate the pre-trained MaST-Pre. Afterward, ablation studies are conducted to analyze the design of our MaST-Pre and show the visualization results.

\subsection{Datasets}
In this paper, to demonstrate the effectiveness of our spatio-temporal representations, we focus on long-term point cloud video tasks, including 4D action recognition and 4D gesture recognition.

\textbf{MSRAction-3D} \cite{msr} and \textbf{NTU-RGBD} \cite{ntu60} are utilized for the action recognition task. 
(1) MSRAction-3D is comprised of 567 videos in 20 daily actions. The average frame number contained in each video is about 40. Following \cite{p4d, fan2022point}, 270 videos are used as the training set and 297 videos are adopted as the test data.
(2) NTU-RGBD consists of 56,880 videos with 60 fine-grained action categories. The frame number of each video is about 30 to 300. Under the cross-subject setting \cite{ntu60}, 40,320 training videos and 16,560 test videos are used.

\textbf{SHREC’17} \cite{de2017shrec} and \textbf{NvGesture} \cite{molchanov2016online} are utilized for the gesture recognition task.
(1) SHREC’17 is comprised of 2800 videos in 28 gestures. Following \cite{de2017shrec}, this dataset is split into 1960 training videos and 840 test videos. 
(2) NvGesture consists of 1532 videos with 25 gesture classes. Following \cite{min2020efficient}, 1050 videos are assigned to the training set and the remaining 482 videos to the test set.

\begin{table}[t]
    \centering
    \small
    \caption{Action recognition accuracy on MSRAction-3D.}
    \setlength{\tabcolsep}{1.3mm}
    \begin{tabular}{l|l|c}
    \toprule
    \multicolumn{2}{c|}{\textbf{Methods}} & \textbf{Accuracy (\%)}\\
    \midrule
    \multirow{8}{*}{{\makecell[l]{Supervised\\Learning}}} &
    MeteorNet~\cite{MeteorNet}    & 88.50 \\
    & PSTNet~\cite{pstnet}        & 91.20 \\
    & PSTNet++~\cite{fan2021deep} & 92.68 \\  
    & Kinet~\cite{zhong2022no}    & 93.27 \\
    & PPTr~\cite{wen2022point}    & 92.33 \\
    & \cellcolor{gray!20}P4Transformer~\cite{p4d}             & \cellcolor{gray!20}90.94 \\
    & \cellcolor{gray!20}PST-Transformer~\cite{fan2022point}  & \cellcolor{gray!20}93.73 \\
    \midrule
    \multirow{2}{*}{\makecell[l]{End-to-end\\Fine-tuning}} 
    & \textbf{P4Transformer \ \ \ \ + MaST-Pre}  & 91.29 \\
    & \textbf{PST-Transformer + MaST-Pre}        & 94.08 \\
    \bottomrule
    \end{tabular}
    \label{MSRAction-3D}
\end{table}

\subsection{Pre-training}
\label{Sec:Pre-training}
During pre-training, given a point cloud video, 24 frames are densely sampled and 1024 points are selected for each frame. Following \cite{p4d, fan2022point}, the frame sampling stride is set to 2/1 on NTU-RGBD/MSRAction-3D and only random scaling is employed for data augmentation. 
For division and embedding, the temporal downsampling rate is set to 2 and the temporal kernel size $l$ of each point tube is set to 3.
Meanwhile, the spatial downsampling rate is set to 32. The radius of each support domain $r$ is set to 0.1/0.3 on NTU-RGBD/MSRAction-3D and the number of neighbor points $n$ within the spherical query is set to 32. 
The masking ratio is set to 75\% unless otherwise specified.

P4Transformer \cite{p4d} is utilized as our encoder, which consists of 10/5 layers of vanilla Transformers on NTU-RGBD/MSRAction-3D.
The decoder is a 4-layer transformer.
To verify the extensibility, PST-Transformer \cite{fan2022point} is also used as an encoder on MSRAction-3D.
Our MaST-Pre is pre-trained for 200 epochs and linear warmup is utilized for the first 10 epochs. 
The AdamW optimizer is used with a batch size of 128, and the initial learning rate is set to 0.001 with a cosine decay strategy. 

\begin{table}[t]
    \centering
    \small
    \caption{Action recognition accuracy (\%) on NTU-RGBD under cross-subject setting.}
    \setlength{\tabcolsep}{1.6mm}
    \begin{tabular}{lc}
    \toprule
    \textbf{Methods}    & \textbf{Acc.} \\
    \midrule
    3DV-Motion~\cite{3dv}                    & 84.5 \\
    3DV-PointNet++~\cite{3dv}                & 88.8 \\
    PSTNet~\cite{pstnet}                     & 90.5 \\
    PSTNet++~\cite{fan2021deep}              & 91.4 \\
    Kinet~\cite{zhong2022no}                 & 92.3 \\
    \rowcolor{gray!20}
    P4Transformer~\cite{p4d}                 & 90.2 \\
    PST-Transformer~\cite{fan2022point}      & 91.0 \\
    \midrule
    \textbf{P4Transformer + MaST-Pre} \small{(End-to-end Fine-tuning)}  & 90.8 \\
    \textbf{P4Transformer + MaST-Pre} \small{(50\% Semi-supervised)} &  87.8 \\
    \bottomrule
    \end{tabular}
    \label{NTU}
\end{table}

\subsection{End-to-end Fine-tuning}
\label{Sec:End-to-end Fine-tuning}
We first evaluate our MaST-Pre by fine-tuning the pre-trained encoder with a new classifier in a supervised manner.
End-to-end fine-tuning experiments are conducted on NTU-RGBD and MSRAction-3D, respectively.
In each experiment, the same dataset is used for pre-training and fine-tuning.

\textbf{MSRAction-3D.} During fine-tuning, 24 frames are densely sampled and 2048 points are selected in each frame. Following~\cite{pstnet}, the spatial search radius is set to 0.7. 
The model is trained for 50 epochs with a batch size of 12 on 4 GPUs. 
We use the AdamW optimizer and the initial learning rate is set to 0.001 with a cosine decay strategy. 
As shown in Table~\ref{MSRAction-3D}, compared with baselines trained in a fully supervised manner, our MaST-Pre introduces accuracy improvements for both P4Transformer and PST-Transformer.
According to prior experience, the masked autoencoder needs to be fed a considerable amount of data in the pre-training stage to learn useful knowledge \cite{he2022masked, feichtenhofer2022masked}. However, the MSRAction-3D dataset is too small to bring significant improvement.

\textbf{NTU-RGBD.} The setup of fine-tuning is the same as pre-training, except that the pre-trained model is fine-tuned for 20 epochs with a batch size of 48 on 8 GPUs and the initial learning rate is set to 0.0005. 
From the end-to-end fine-tuning in Table~\ref{NTU}, we can see that our pre-training method introduces an accuracy improvement compared with the baseline. By predicting the spatio-temporal structure, our MaST-Pre learns appearance and motion information during pre-training.

\subsection{Semi-supervised Learning}
\label{Sec:Semi-supervised Learning}
We also evaluate the learned representations using a semi-supervised learning experiment. 
Specifically, the cross-subject training set of NTU-RGBD is used for pre-training, and then only a 50\% training set is used for fine-tuning in a supervised manner. 
The setup of our semi-supervised learning experiment is the same as end-to-end fine-tuning on NTU-RGBD (Sec. \ref{Sec:End-to-end Fine-tuning}).

From Table~\ref{NTU}, we can see that the 50\% semi-supervised result produced by our MaST-Pre achieves comparable performance to the fully supervised baseline even with only limited annotated data. This clearly demonstrates that MaST-Pre learns high-quality representations.

\begin{table}[t]
    \centering
    \small
    \caption{Gesture recognition accuracy (\%) on NvGesture (NvG) and SHREC'17 (SHR).}
    \setlength{\tabcolsep}{2.5mm}
    \begin{tabular}{lcc}
    \toprule
    \textbf{Methods} &\textbf{NvG} &\textbf{SHR}\\
    \midrule
    FlickerNet~\cite{flickernet}           & 86.3    & -    \\
    PLSTM~\cite{min2020efficient}     & 85.9    & 87.6 \\
    PLSTM-PSS~\cite{min2020efficient}      & 87.3    & 93.1 \\
    Kinet~\cite{zhong2022no}               & 89.1    & 95.2 \\
    \rowcolor{gray!20}
    P4Transformer~\cite{p4d} \ (30 Epochs)   & 84.8   & 87.5\\
    \rowcolor{gray!20}
    P4Transformer~\cite{p4d} \ (50 Epochs)   & 87.7   & 91.2 \\
    \midrule
    \textbf{P4Transformer + MaST-Pre} \ (30 Epochs)  & 87.6 & 90.2 \\
    \textbf{P4Transformer + MaST-Pre} \ (50 Epochs)  & 89.3 & 92.4 \\
    \bottomrule
    \end{tabular}
    \label{NvGesture}
\end{table}

\begin{figure*}[ht]
	\centering
	\includegraphics[width=1\linewidth]{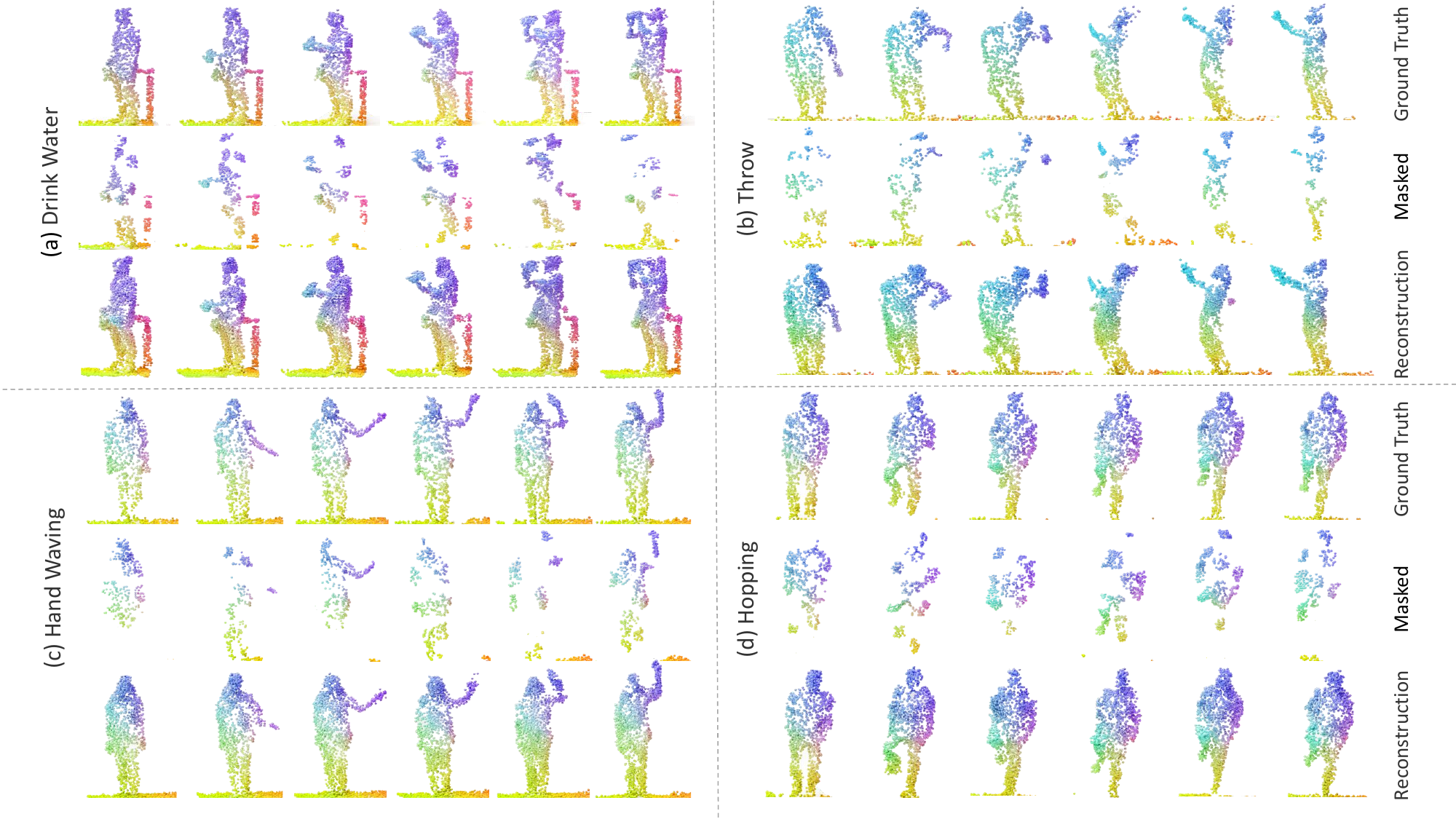}	\caption{Visualization of reconstruction results. For each action sample, the ground truth lies in the first row, the masked video lies in the second row, and the result lies in the third row.}
	\label{fig4}
\end{figure*}

\subsection{Transfer Learning}
\label{Sec:Transfer Learning}
To evaluate the generalization ability of the representations learned by MaST-Pre, we conduct experiments by transferring the pre-trained encoder to other datasets. 
Specifically, the encoder is first pre-trained on NTU-RGBD following the setup in Sec.~\ref{Sec:Pre-training}, and then fine-tuned with a new classifier on NvGesture and SHREC’17, respectively. 
Our transfer experiments are not only cross-dataset but also cross-task, \ie, from action recognition to gesture recognition. 
We compare our fine-tuned results to the fully supervised baseline in Table~\ref{NvGesture}.

During fine-tuning, an AdamW optimizer with a batch size of 24 is used, and the initial learning rate is set to 0.002 with a cosine decay strategy. 
The pre-trained model is fine-tuned for 50 epochs on NvGesture and SHREC’17.
As shown in Table~\ref{NvGesture}, our MaST-Pre pre-training facilitates the P4Transformer to produce superior accuracy compared to the fully supervised baseline. 
Moreover, our MaST-Pre also performs faster convergence. Compared with the baseline without pre-training, significant improvements are achieved after fine-tuning for only 30 epochs (\eg, 84.8\% $\rightarrow$ 87.6\% on NvGesture and 87.5\% $\rightarrow$ 90.2\% on SHREC’17). 
This demonstrates that our MaST-Pre has excellent generalization ability across different tasks, facilitating the accuracy improvement of downstream tasks.

\subsection{Ablation Studies}
\label{Sec:Ablation Studies}
In order to balance authority and efficiency, the experiments of ablation studies are conducted on 10\% NTU-RGBD, which contains 4032 training videos and 1656 test videos with category balance.

\begin{table}[t]
    \centering
    \small
    \setlength{\tabcolsep}{1.5mm}
    \caption{Ablation studies on pre-training architectures.}
    \begin{tabular}{lcc|c}
    \toprule 
    & \textbf{Appearance Stream} & \textbf{Motion Stream}  & \textbf{Acc. (\%)} \\
    \midrule
    \rowcolor{gray!20}\footnotesize{A0}        &           &              & 65.85  \\
    \footnotesize{A1}        & $\checkmark$ &              &  70.13  \\
    \footnotesize{A2 (Ours)} & $\checkmark$ & $\checkmark$ & \textbf{78.25} \\
    \bottomrule
    \end{tabular}
    \label{architecture}
\end{table}

\textbf{Architecture Design.} 
Our MaST-Pre utilizes a two-stream prediction to jointly learn both appearance and motion information. To demonstrate the effectiveness of this architecture, we first present the performance of model A0 as the 10\% NTU-RGBD baseline in a fully supervised manner (65.85\%). Then, model A1 is developed by removing the motion prediction stream. Quantitative results are presented in Table~\ref{architecture}. It shows that with solely the appearance stream, the performance gain introduced by model A1 pre-training is limited. When these two streams are combined, comprehensive information can be learned and superior accuracy is achieved by our model A2 (65.85\% $\rightarrow$ 78.25\%). This demonstrates the effectiveness of our mask-based pre-training on point cloud videos and the necessity of the task of explicitly predicting motions.

\begin{table}[t]
    \centering
    \small
    \caption{Ablation studies on masking ratios.}
    \setlength{\tabcolsep}{4.2mm}
    \begin{tabular}{l|cccc}
    \toprule 
      & \footnotesize{B1} & \footnotesize{B2 (Ours)} & \footnotesize{B3} \\
    \midrule
    \textbf{Masking Ratio}     & 65\% & 75\%  & 85\%   \\
    \textbf{Accuracy (\%)}  & \ 76.91 \ & \ \textbf{78.25} \ & \ 77.20 \ \\
    \bottomrule
    \end{tabular}
    \label{masking ratio}
\end{table}


\textbf{Masking Ratio.} 
The masking operation plays a critical role in our MaST-Pre. Therefore, we conduct experiments to study different masking ratios, and the results are presented in Table~\ref{masking ratio}. 
It shows that a high masking ratio is beneficial to our MaST-Pre and the highest accuracy is achieved at 75\% masking ratio (model B2).

\begin{table}[t]
    \centering
    \small
    \caption{Ablation studies on the appearance stream.}
    \setlength{\tabcolsep}{1.5mm}
    \begin{tabular}{lc|c|c}
    \toprule 
    & \textbf{\# Frames} & \textbf{Temporal \& Spatial} & \textbf{Accuracy (\%)} \\
    \midrule
    \footnotesize{D1}        & 1 & -          &  77.34 \\
    \midrule
    \footnotesize{D2}        & 3 & Coupling   &  58.60\\
    \footnotesize{D3 (Ours)} & 3 & Decoupling & \textbf{78.25} \\
    \bottomrule
    \end{tabular}
    \label{appearance}
\end{table}

\begin{figure*}[ht]
	\centering
	\includegraphics[width=1\linewidth]{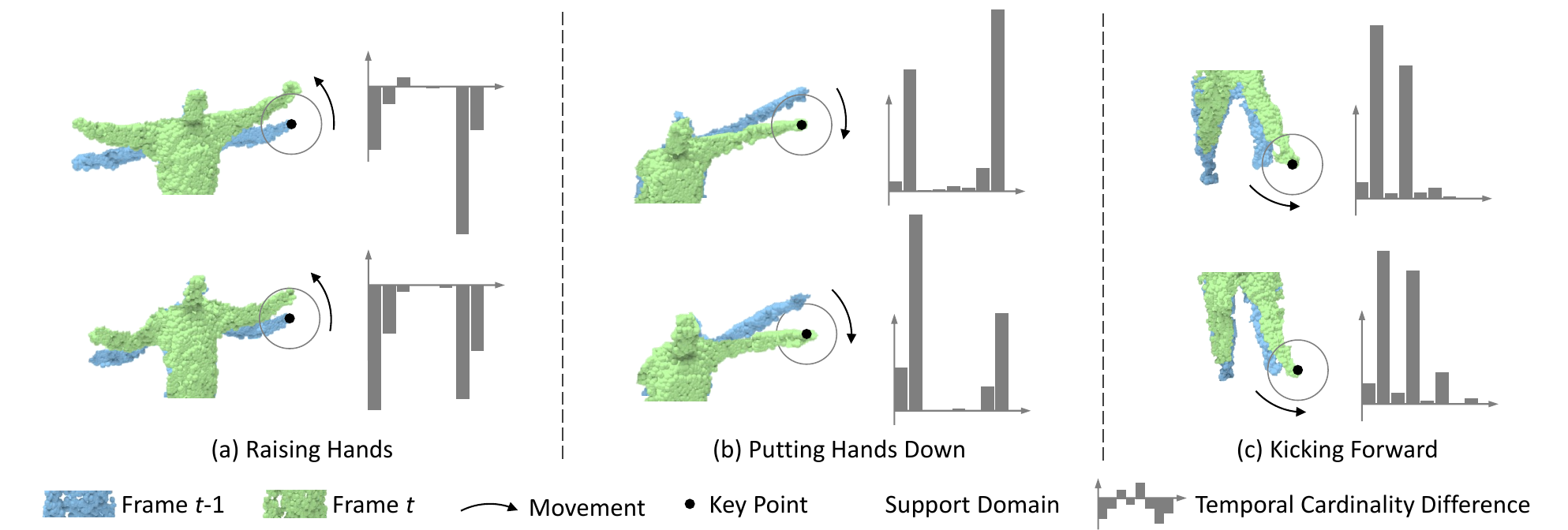}
	\caption{Samples of temporal cardinality difference, computed by subtracting cardinality histograms of frame $t\!-\!1$ from frame $t$.}
	\label{fig5}
\end{figure*}

\begin{table*}[ht]
    \caption{Ablation studies on the temporal cardinality difference.}
    \hspace{0.4mm}
    \begin{minipage}[t]{0.35\linewidth}
        \centering
        \small
        \resizebox{\textwidth}{!}{
        \begin{tabular}{l|cccc}
        \toprule 
          & \footnotesize{E1} & \footnotesize{E2 (Ours)} & \footnotesize{E3}  \\
        \midrule
        \textbf{\# Section}     & 4     & 8              & 16 \\
        \textbf{Accuracy (\%)}  & 77.85 & \textbf{78.25} & 69.50 \\
        \bottomrule
        \end{tabular}
        }
        \label{Section}
    \end{minipage}
    \hspace{0.5mm}
    \ 
    \begin{minipage}[t]{0.29\linewidth}
        \centering
        \small
        \resizebox{\textwidth}{!}{
        \begin{tabular}{l|cc}
        \toprule 
                               & \footnotesize{F1} & \footnotesize{F2 (Ours)}  \\
        \midrule
        \textbf{Interpolation}    &  \ding{55}        & \ding{51} \\
        \textbf{Accuracy (\%)} &  76.69       &  \textbf{78.25}  \\
        \bottomrule
        \end{tabular}
        }
        \label{Interpolation}
    \end{minipage}
    \hspace{0.5mm}
    \
    \begin{minipage}[t]{0.29\linewidth}
        \centering
        \small
        \resizebox{\textwidth}{!}{
        \begin{tabular}{l|cc}
        \toprule 
        & \footnotesize{G1 (Ours)} & \footnotesize{G2 }  \\
        \midrule
        \textbf{\# Stride}     & 1                 & 2 \\
        \textbf{Accuracy (\%)} & \textbf{78.25}    & 75.85 \\
        \bottomrule
        \end{tabular}
        }
        \label{Frame Stride}
    \end{minipage}
\label{temporal cardinality difference}
\end{table*}


\textbf{Appearance Stream.}
Right reconstruction targets in the appearance stream contribute to the performance of our MaST-Pre. 
For model D3, in each point tube, the reconstruction loss (Eq.~\ref{app-loss}) is first calculated in each frame separately and then aggregated over $l$ frames, which is a decoupled manner.
In contrast, model D2 calculates the reconstruction loss in a coupled manner by considering all points together. 
We also develop model D1 to reconstruct only the middle frame of each point tube.

As shown in Table~\ref{appearance}, model D2 brings no accuracy gain and is even worse than the baseline model A0 (58.60\% vs. 65.85\%). In addition, D3 outperforms D1 and D2. This is because D3 implicitly learns spatio-temporal information during the process of reconstructing decoupled point tubes. We further visualize the reconstruction results in Fig.~\ref{fig4}.

\textbf{Temporal Cardinality Difference.}
In order to investigate the temporal cardinality difference, we present the results under different sections, interpolations, and strides in Table~\ref{temporal cardinality difference}.
We develop the model E1/E3 to divide the support domain into 4/16 sections, while their accuracy is lower than E2 with 8 sections. This is because small space resolution will introduce noises and large resolution makes temporal cardinality difference insensitive to motions.

Next, we develop model F1 by removing interpolation. While F2 outperforms F1 because interpolation improves its robustness. Finally, we develop model G2 to calculate the cardinality difference with temporal stride 2, but its performance is worse than G1 with stride 1. This is because a large temporal stride cannot capture fine-grained motions.

We further visualize multiple samples of temporal cardinality difference in Fig.~\ref{fig5} to demonstrate its effectiveness in modeling motions.
We present three typical actions, each consisting of two samples. 
As shown in Fig.~\ref{fig5}(a), the temporal cardinality differences of the two \textit{raising hands} actions project extremely similar motion patterns.
Points in the first and seventh octants flow out heavily over time. 
Meanwhile, temporal cardinality differences within \textit{putting hands down} (Fig.~\ref{fig5}(b)) also display similarities, as well as in \textit{kicking forward} (Fig.~\ref{fig5}(c)). 
In particular, the temporal cardinality differences between \textit{raising hands} and \textit{putting hands down} are approximately reversible, which reflects its effectiveness in modeling dynamics.

\begin{table}[t]
    \centering
    \small
    \caption{Time (mins/epoch) and memory (MiB) complexities.}
    \setlength{\tabcolsep}{0.5mm}
    \begin{tabular}{ll|c|c|c|c}
    \toprule 
    & \multicolumn{1}{l|}{\textbf{Architectures}} & \textbf{Encoder} & \textbf{Time} & \textbf{Memory} & \textbf{Acc. (\%)} \\
    \midrule
    \footnotesize{H1}        & Only Appearance & w/o \footnotesize{$\mathtt{[M]}$}  & 5.4 & 6414 & 70.13 \\
    \footnotesize{H2 (Ours)} & Two Streams  & w/o \footnotesize{$\mathtt{[M]}$}     & 6.2 & 6418 &  \textbf{78.25}\\
    \bottomrule
    \end{tabular}
    \label{complexities}
\end{table}

\textbf{Computational Complexity.}
Table~\ref{complexities} shows the pre-training complexity and corresponding fine-tuning accuracy of the two models.
After adding the motion prediction stream, model H2 achieves much higher accuracy than H1 with only a minor increase in pre-training complexities (70.13\% $\rightarrow$ 78.25\%). 

\section{Conclusion}
In this paper, we introduce a masked spatio-temporal structure prediction method for point cloud video pre-training, termed as MaST-Pre. 
For modeling subtle dynamics, the temporal cardinality difference is proposed, which can be calculated online directly from inputs. 
Based on point-tube masking, MaST-Pre jointly conducts point cloud video reconstruction and temporal cardinality difference prediction to learn both appearance and motion information.
Experiments on four benchmarks show that our MaST-Pre is an effective pre-training framework to boost the performance of point cloud video understanding.

\noindent\textbf{Acknowledgments.} This work was partially supported by the Fundamental Research Funds for the Central Universities (No. 226-2023-00048), the National Natural Science Foundation of China (No. U20A20185, 61972435), the Guangdong Basic and Applied Basic Research Foundation (2022B1515020103), the Shenzhen Science and Technology Program (No. RCYX20200714114641140), and the Chongqing Technology Innovation and Application Development Special Key Project(cstc2021jscx-cylhX0006).

{\small
\bibliographystyle{ieee_fullname}
\bibliography{egbib}
}

\end{document}